\documentclass[10pt,twocolumn,letterpaper]{article}

\usepackage{3dv}
\usepackage{times}
\usepackage{epsfig}
\usepackage{graphicx}
\usepackage{amsmath}
\usepackage{amssymb}

\usepackage{multirow}
\usepackage{bm}
\usepackage{placeins} % For FloatBarrier
\usepackage{flafter}  % For FloatBarrier
\usepackage{balance}
\usepackage{lipsum}
\usepackage{enumitem}
\setlist[itemize]{noitemsep, topsep=3pt, leftmargin=*}
\setlist[enumerate]{noitemsep, topsep=3pt}

\usepackage[linesnumbered,ruled,vlined]{algorithm2e}

\SetCommentSty{mycommfont}
\SetKwInput{KwInput}{Input}                
\SetKwInput{KwOutput}{Output}     
\SetKw{Continue}{continue}

\usepackage{setspace}

\newlength{\textfloatsepsave} 
\newlength{\floatsepsave} 

\newdimen\origiwspc
\newdimen\origiwstr

\newenvironment{fontspace}[2]
{\par
    \origiwspc=\fontdimen2\font% original inter word space
    \origiwstr=\fontdimen3\font% original inter word stretch
    \fontdimen2\font=#1\origiwspc
    \fontdimen3\font=#2\origiwstr
}
{\par
    \fontdimen2\font=\origiwspc
    \fontdimen3\font=\origiwstr
}

\usepackage[pagebackref=true,breaklinks=true,letterpaper=true,colorlinks,bookmarks=false]{hyperref}

\threedvfinalcopy % *** Uncomment this line for the final submission

\def\threedvPaperID{27} % *** Enter the 3DV Paper ID here

\begin{document}

%%%%%%%%% TITLE
\title{Robust Uncertainty-Aware Multiview Triangulation}

\author{Seong Hun Lee \hspace{25pt} Javier Civera\thanks{This work was partially supported by the Spanish govt. (PGC2018- 096367-B-I00) and the Arag{\'{o}}n regional govt. (DGA-T45{\_}17R/FSE).} \\
I3A, University of Zaragoza, Spain\\
{\tt\small \{seonghunlee, jcivera\}@unizar.es}
}

\maketitle
% \thispagestyle{empty}

%%%%%%%%% ABSTRACT
\begin{abstract}
\vspace{-1em}
   We propose a robust and efficient method for multiview triangulation and uncertainty estimation.
   Our contribution is threefold:
   First, we propose an outlier rejection scheme using two-view RANSAC with the midpoint method.
   By prescreening the two-view samples prior to triangulation, we achieve the state-of-the-art efficiency.
   Second, we compare different local optimization methods for refining the initial solution and the inlier set.
   With an iterative update of the inlier set, we show that the optimization provides significant improvement in accuracy and robustness.
   Third, we model the uncertainty of a triangulated point as a function of three factors:
   the number of cameras, the mean reprojection error and the maximum parallax angle.
   Learning this model allows us to quickly interpolate the uncertainty at test time.
   We validate our method through an extensive evaluation.
   %Through an extensive evaluation, we validate our method and provide important insight into multiview triangulation.
\end{abstract}

%%%%%%%%% BODY TEXT
\vspace{-1em}
\section{Introduction}
\vspace{-0.2em}
Multiview triangulation refers to the problem of locating the 3D point given its projections in multiple views of known calibration and pose.
It plays a fundamental role in many applications of computer vision, \eg, structure-from-motion \cite{building_rome, moulon2013global,sfm_revisited}, visual(-inertial) odometry \cite{svo2,ptam, okvis} and simultaneous localization and mapping \cite{dt_slam, orb-slam2, vins_mono}.

Under the assumption of perfect information (\ie, image measurements, calibration and pose data without noise and outliers), triangulation simply amounts to intersecting the backprojected rays corresponding to the same point.
In practice, however, noise and outliers are often inevitable.
This makes the triangulation problem nontrivial.
From a practical perspective, the following aspects should be taken into account when considering a triangulation method:

\textbf{1. Is it applicable to multiple views?}
Some methods are developed specifically for two or three views (\eg, two-view optimal methods \cite{hartley_triangulation, kanatani, closed_form_optimal_triangulation_based_angular_errors, lindstrom, closed_form_oliensis}, two-view midpoint methods \cite{beardsley_midpoint2, hartley_triangulation, triangulation_why_optimize}, three-view optimal methods \cite{martin_three_view, hedborg_three_view, kukelova_three_view,stewenius_three_view}).
For more than two or three views, these methods are not directly applicable unless they are incorporated in, for example, a RANSAC \cite{ransac} framework.

\textbf{2. Is it robust to outliers?}
Many existing multiview triangulation methods are sensitive to outliers, \eg, the linear methods \cite{hartley_triangulation}, the midpoint-based methods \cite{ramalingam2006generic,yang2019iteratively}, the $L_2$-optimal methods \cite{aholt2012qcqp, kahl2008practical,kahl2007globally, lu2007fast} and the $L_\infty$-optimal methods \cite{agarwal2008fast,dai2012novel, hartley2004Linf,kahl2008multiple}.
To deal with outliers, various methods have been proposed, \eg, outlier rejection using the $L_\infty$ norm \cite{li2007practical,olsson2010outlier, sim2006removing}, $k$-th median minimization \cite{ke2007quasiconvex} and iteratively reweighted least squares \cite{aftab2015convergence}.
We refer to \cite{L2_kang} for a comprehensive review of the robust triangulation methods.

\textbf{3. Is it fast?}
While the aforementioned methods can handle a moderate number of outliers, they either fail or incur excessive computational cost at high outlier ratios \cite{sfm_revisited}. 
For this reason, RANSAC is often recommended as a preprocessing step \cite{li2007practical,olsson2010outlier,sim2006removing}.
In \cite{sfm_revisited}, an efficient method using two-view RANSAC is proposed.
%This method provides a good balance between the robustness and speed, which is important for structure-from-motion in practice.

\textbf{4. Does it estimate the uncertainty?}
To our knowledge, none of the aforementioned works provide the uncertainty estimate for the triangulated point.
Knowing this uncertainty can be useful for point cloud denoising \cite{wolff2016point}, robust localization \cite{sattler2011fast,svarm2014accurate} and mapping \cite{dpptam, semidense_orbslam}, among others.

%\vspace*{\fill}

In this work, we propose a robust and efficient method for uncertainty-aware multiview triangulation.
Our contributions are summarized as follows:
\begin{enumerate}[leftmargin=1.5em]
    \item 
    In Sect.\hspace{0.5em}\ref{subsec:two_view_ransac}, we propose an outlier rejection scheme using two-view RANSAC with the midpoint method.
    By reformulating the midpoint, we screen out the bad samples even before computing the midpoint.
    This improves the efficiency when the outlier ratio is high.
    
    \vspace{0.3em}
    
    \item 
    In Sect.\hspace{0.5em}\ref{subsec:local_optimization}, we revisit three existing local optimization methods, one of which is the Gauss-Newton method.
    For this method, we present an efficient computation of the Jacobian matrix.
    We closely evaluate the three methods with an iterative update of the inlier set.
    
    \vspace{0.3em}
   
    \item 
    In Sect.\hspace{0.5em}\ref{subsec:uncertainty}, we model the uncertainty of a triangulated point as a function of three factors: the number of (inlying) cameras, the mean reprojection error and the maximum parallax angle.
    We propose to learn this model from extensive simulations, so that at test time, we can estimate the uncertainty by interpolation.
    The estimated uncertainty can be used to control the 3D accuracy.
\end{enumerate}

\newpage
\clearpage
\begin{algorithm}[t]
\setstretch{0.97}
\label{al:proposed}
\caption{Proposed Multiview Triangulation}
\DontPrintSemicolon
\KwInput{$\mathcal{V}$ and $\mathbf{u}_i$, $\mathbf{K}_i$, $\mathbf{R}_i$, $\mathbf{t}_i$ for all $i\in\mathcal{V}$, \\ \hspace{3em}$\eta$, $\delta_{2D}$, $\delta_\mathrm{epipolar}$, $\delta_\mathrm{lower}$, $\delta_\mathrm{upper}$, $\delta_\mathrm{update}$, $\delta_\mathrm{pair}.\hspace{-5em}$}
\KwOutput{$\mathbf{x}^w_\mathrm{est}$, $\mathcal{I}$, $\overline{e}_{2D}$, $\sigma_{3D}$.}

\tcc{Initialization}

$\mathbf{x}^w_\mathrm{est}\gets\mathbf{0}$; \ \ 
$\mathcal{I}\gets\{\}$; \ \ 
$\overline{e}_{2D}\gets \infty$; \ \ 
$\sigma_{3D}\gets \infty$; \ \ 

$\widehat{\mathbf{f}}^w_i\gets\mathbf{0}$, \ 
$\mathbf{c}_i^w \gets -\mathbf{R}_i^\top \mathbf{t}_i$, \ 
$\mathbf{P}_i \gets \left[\mathbf{R}_i \ | \ \mathbf{t}_i\right]$ for all $i \in\mathcal{V};\hspace{-5em}$

compute $\mathbf{M}_1, \cdots, \mathbf{M}_5$ using \eqref{eq:M1}--\eqref{eq:M5};

compute $\mathbf{b}_{1i}, \cdots, \mathbf{b}_{6i}$ for all $i\in\mathcal{V}$ using \eqref{eq:b1}--\eqref{eq:b6};

compute $\mathbf{a}_{1i}, \cdots, \mathbf{a}_{6i}$ for all $i\in\mathcal{V}$ using \eqref{eq:a1}--\eqref{eq:a6};

compute $\mathbf{A}_i$ for all $i\in\mathcal{V}$ using \eqref{eq:A};

\tcc{(1) Two-view RANSAC (Sect.\ref{subsec:two_view_ransac})\hspace{-10em}}

$m_\mathrm{min}\gets n(n-1)/2$; \quad
$C_\mathrm{min}\gets\infty$;

\While{$m < m_\mathrm{min}$}
{

$m\gets m+1$;

Pick a random pair of views $j,k\in\mathcal{V}$;

Perform Alg. \ref{al:midpoint} for $(j,k)$.

\lIf{$b_\mathrm{good}=\mathrm{false}$}{\Continue;}

$\mathbf{x}^w_\mathrm{est}\gets\mathbf{x}^w_\mathrm{mid}$;

compute $\mathbf{M}_6$, $\mathbf{M}_7$, $\mathbf{M}_8$ using \eqref{eq:M6}--\eqref{eq:M8};

compute $\mathbf{e}_{2D}$, $\mathcal{I}$, $C$ using \eqref{eq:reproj_error_fast}, \eqref{eq:inlier_set} and \eqref{eq:cost};

\lIf{$C\geq C_\mathrm{min}$}{\Continue;}

$C_\mathrm{min}{\gets\hspace{0.2em}}C$; \ 
$\mathbf{x}^{w*}_\mathrm{est}{\gets\hspace{0.2em}}\mathbf{x}^w_\mathrm{est}$; \ 
$\mathcal{I}^*{\gets\hspace{0.2em}}\mathcal{I}$; \ 
$\mathbf{M}^*_{6,7,8}{\gets\hspace{0.2em}}\mathbf{M}_{6,7,8};\hspace{-5em}$

$\epsilon\gets\max(|\mathcal{I^*}|,2)/n$; \ \ 
$m_\mathrm{min}\gets\displaystyle\frac{\log\left(1-\eta\right)}{\log\left(1-\epsilon^2\right)}$; \label{line:adaptive_stopping}

}

\lIf{$C_\mathrm{min}=\infty$}{go to Line \ref{line:multiview_return};}

\tcc{(2) Local optimization (Sect.\ref{subsec:local_optimization})\hspace{-10em}}

$\mathbf{x}^w_\mathrm{est}\gets\mathbf{x}^{w*}_\mathrm{est}$; \ \ 
$\mathcal{I}\gets\mathcal{I}^*$; \ \ 
$\mathbf{M}_{6,7,8}\gets\mathbf{M}^*_{6,7,8}$;

Perform Alg. \ref{al:gn};

\tcc{(3) Uncertainty estimation (Sect.\ref{subsec:uncertainty})\hspace{-10em}}

Perform Alg. \ref{al:uncertainty};

\Return $\mathbf{x}^w_\mathrm{est}$, $\mathcal{I}$, $\overline{e}_{2D}$, $\sigma_{3D}$; \label{line:multiview_return}
\end{algorithm}

\newpage
\setlength{\textfloatsep}{1em}
The proposed approach is detailed in Alg. \ref{al:proposed}.
See the supplementary material for the nomenclature.
To download our code, go to \url{http://seonghun-lee.github.io}.

\section{Preliminaries and Notation}
\vspace{-0.5em}
We use bold lowercase letters for vectors, bold uppercase letters for matrices, and light letters for scalars.
We denote the Hadamard product, division and square root by ${\mathbf{A}\circ\mathbf{B}}$, ${\mathbf{A}\oslash\mathbf{B}}$ and ${\mathbf{A}^{\circ1/2}}$, respectively.
The Euclidean norm of a vector $\mathbf{v}$ is denoted by $\lVert\mathbf{v}\rVert$, and the unit vector by $\widehat{\mathbf{v}}=\mathbf{v}/\lVert\mathbf{v}\rVert$.
The angle between two vectors $\mathbf{a}$ and $\mathbf{b}$ is denoted by $\angle(\mathbf{a},\mathbf{b})\in[0,\pi/2]$.
We denote the vectorization of an $n\times m$ matrix by $\mathrm{vec}(\cdot)$ and its inverse by $\mathrm{vec}^{-1}_{n\times m}(\cdot)$.

Consider a 3D point $\mathbf{x}^w=[x^w, y^w, z^w]^\top$ in the world reference frame and a perspective camera $c_i$ observing this point.
In the camera reference frame, the 3D point is given by
$\mathbf{x}_i=[x_i, y_i, z_i]^\top=\mathbf{R}_i\mathbf{x}^w+\mathbf{t}_i=\mathbf{P}_i\widetilde{\mathbf{x}}^w$,
where $\mathbf{R}_i$ and $\mathbf{t}_i$ are the rotation and translation that relate the reference frame $c_i$ to the world, $\mathbf{P}_i=\left[\mathbf{R}_i \ | \ \mathbf{t}_i\right]$ is the extrinsic matrix, and $\widetilde{\mathbf{x}}^w=\left[x^w, y^w, z^w, 1\right]^\top$ is the homogeneous coordinates of $\mathbf{x}^w$.
In the world frame, the camera position is given by $\mathbf{c}^w_i=-\mathbf{R}_i^\top\mathbf{t}_i$.
Let $\widetilde{\mathbf{u}}_i=\left[\mathbf{u}_i^\top, 1\right]^\top=[u_i, v_i, 1]^\top$ be the homogeneous pixel coordinates of the point and $\mathbf{K}_i$ the camera calibration matrix.
Then, the normalized image coordinates $\mathbf{f}_i=[x_i/z_i, y_i/z_i,1]^\top$ are obtained by $\mathbf{f}_i=\mathbf{K}_i^{-1}\widetilde{\mathbf{u}}_i$.

Let $\mathcal{V}=\{1,2,\cdots,n\}$ be the set of all views in which the point is observed.
The aim of multiview triangulation is to find the best estimate of $\mathbf{x}^w$ given that noisy $\mathbf{u}_i$, $\mathbf{R}_i$, $\mathbf{t}_i$ and $\mathbf{K}_i$ are known for all $i\in\mathcal{V}$.
Once we have the estimate $\mathbf{x}^w_\mathrm{est}$, the 3D error is given by $e_{3D}=\lVert\mathbf{x}^w_\mathrm{est}-\mathbf{x}^w\rVert$, and the 2D error (aka the reprojection error) is given by
\vspace{-0.5em}
\begin{flalign}
    &\mathbf{e}_{2D}=\begin{bmatrix}
    \lVert\mathbf{u}_1-\mathbf{u}'_1\rVert, \ \lVert\mathbf{u}_2-\mathbf{u}'_2\rVert, \ \cdots, \ \lVert\mathbf{u}_n-\mathbf{u}'_n\rVert
    \end{bmatrix}^\top, \hspace{-2em} \label{eq:reproj_error}& 
\end{flalign}
\vspace{-1.5em}

\noindent
where $\mathbf{u}'_i$ is the reprojection of $\mathbf{x}^w_\mathrm{est}$ in $c_i$.
To compute $\mathbf{e}_{2D}$ compactly, we define the following matrices:
\vspace{-0.5em}
\begin{align}
    \mathbf{M}_1 &:= \left[k_{113}-u_1, \cdots, k_{n13}-u_n \right], \label{eq:M1}\\
    \mathbf{M}_2 &:= \left[k_{123}-v_1, \cdots, k_{n23}-v_n \right],\label{eq:M2}\\
    \begin{split}
    \mathbf{M}_3 &:=\left[\left(k_{111}(\mathbf{P}_1)_\mathrm{row1}^\top+k_{112}(\mathbf{P}_1)_\mathrm{row2}^\top\right), \cdots,\right.\\ &\quad \ \ \  \left.\left(k_{n11}(\mathbf{P}_n)_\mathrm{row1}^\top+k_{n12}(\mathbf{P}_n)_\mathrm{row2}^\top\right)\right],
    \end{split} \label{eq:M3}
    \\
    \begin{split}
    \mathbf{M}_4 &:=\left[\left(k_{121}(\mathbf{P}_1)_\mathrm{row1}^\top+k_{122}(\mathbf{P}_1)_\mathrm{row2}^\top\right), \cdots,\right.\\ &\quad \ \ \  \left.\left(k_{n21}(\mathbf{P}_n)_\mathrm{row1}^\top+k_{n22}(\mathbf{P}_n)_\mathrm{row2}^\top\right)\right],
    \end{split}
    \\
    \mathbf{M}_5 &:= \left[(\mathbf{P}_1)_\mathrm{row3}^\top, \cdots, \mathbf{(P}_n)_\mathrm{row3}^\top\right], \label{eq:M5}\\
    \mathbf{M}_6 &:= \left(\widetilde{\mathbf{x}}^w_\mathrm{est}\right)^\top\mathbf{M}_5, \label{eq:M6}
    \\
    \mathbf{M}_7 &:= \mathbf{M}_1+\left(\left(\widetilde{\mathbf{x}}^w_\mathrm{est}\right)^\top\mathbf{M}_3\right)\oslash\mathbf{M}_6,\\
    \mathbf{M}_8 &:= \mathbf{M}_2+\left(\left(\widetilde{\mathbf{x}}^w_\mathrm{est}\right)^\top\mathbf{M}_4\right)\oslash\mathbf{M}_6. \label{eq:M8}
\end{align}
\vspace{-1.5em}

\noindent
where $k_{ijk}$ is the element of $\mathbf{K}_i$ at the $j$-th row and $k$-th column.
Then, $\mathbf{e}_{2D}$ can be obtained as follows:
\vspace{-0.5em}
\begin{equation}
\label{eq:reproj_error_fast}
    \mathbf{e}_{2D}^\top = \left(\mathbf{M}_7\circ\mathbf{M}_7 +\mathbf{M}_8\circ\mathbf{M}_8\right)^{\circ1/2}.
\vspace{-0.5em}
\end{equation}
We provide the derivation in the supplementary material.
Note that $\mathbf{M}_3$, $\mathbf{M}_4$ and $\mathbf{M}_5$ are independent of the point, and thus can be precomputed for efficiency.
%Therefore, by using \eqref{eq:reproj_error_fast} instead of \eqref{eq:reproj_error}, we can reduce the number of operations when we reproject a point iteratively or triangulate multiple points. 

\begin{figure}[t]
\vspace{-0.5em}
 \centering
 \includegraphics[width=0.28\textwidth]{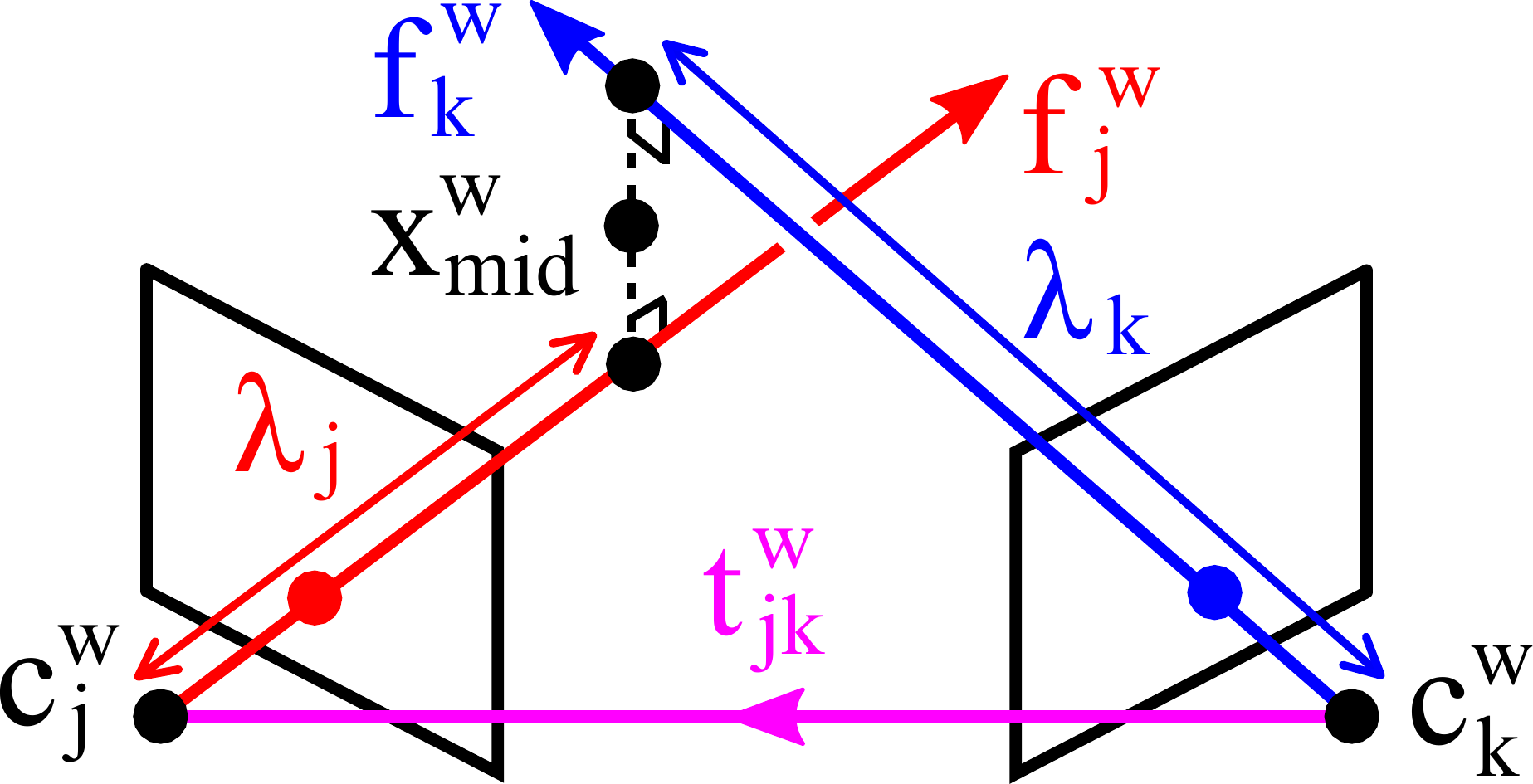}
\caption{The midpoint of the two corresponding rays.
 }
\label{fig:midpoint}
\end{figure}

We define the positive z-axis of the camera as the forward direction.
This means that if the $i$-th element of $\mathbf{M}_6$ is negative, the point is behind the camera $c_i$, violating the cheirality \cite{hartley_book}.
Hence, given the estimated point $\mathbf{x}^w_\mathrm{est}$, the corresponding set of inliers is obtained by
\vspace{-0.1em}
\begin{equation}
\label{eq:inlier_set}
    \mathcal{I}=\{i\in\mathcal{V} \ | \ \left(\mathbf{e}_{2D}\right)_i < \delta_{2D} \land (\mathbf{M}_6)_i > 0 \},
\vspace{-0.1em}
\end{equation}
where $(\cdot)_i$ indicates the $i$-th element, and $\delta_{2D}$ is the inlier threshold.
We denote the number of elements of $\mathcal{I}$ by $|\mathcal{I}|$.
We define the maximum parallax angle of $\mathcal{I}$ as follows:
\vspace{-0.1em}
\begin{align}
    \beta_\text{max}:=&\max\left\{\angle\left(\mathbf{f}^w_j, \mathbf{f}^w_k\right) \ | \ j,k\in\mathcal{I}\right\}\label{eq:beta_max}\\
    =&\cos^{-1}\left(\min\left\{\left|\widehat{\mathbf{f}}^w_j\cdot\widehat{\mathbf{f}}_k^w\right| \ | \ j,k\in\mathcal{I}\right\}\right),\label{eq:beta_max2}
\end{align}

\vspace{-0.5em}

\noindent
where $\widehat{\mathbf{f}}^w_j$ and $\widehat{\mathbf{f}}^w_k$ are the two corresponding unit rays from camera $j$ and $k$ expressed in the world frame, \ie,
\begin{flalign}
    &\widehat{\mathbf{f}}^w_j {=\hspace{0.1em}}\mathbf{R}_j^{\hspace{-0.2em}\top}\widehat{\mathbf{f}}_j, \  
    \widehat{\mathbf{f}}^w_k {=\hspace{0.1em}}\mathbf{R}_k^{\hspace{-0.2em}\top}\widehat{\mathbf{f}}_k \hspace{0.5em}\text{with}\hspace{0.5em}   
    \mathbf{f}_j{=\hspace{0.1em}}\mathbf{K}_j^{\text{--}1}\widetilde{\mathbf{u}}_j,  \ 
    \mathbf{f}_k{=\hspace{0.1em}}\mathbf{K}_k^{\text{--}1}\widetilde{\mathbf{u}}_k.\hspace{-5em}&\label{eq:world_ray}
\end{flalign}

\section{Method}
\subsection{Fast Two-View RANSAC for Outlier Rejection}
\label{subsec:two_view_ransac}

\begin{algorithm}[t]
\setstretch{0.9}
\caption{Midpoint Method with Early \mbox{Termination\hspace{-5em}}}
\label{al:midpoint}
\DontPrintSemicolon
\KwInput{$\mathbf{u}_j$, $\mathbf{u}_k$, $\mathbf{K}_j$, $\mathbf{K}_k$, $\mathbf{R}_j$, $\mathbf{R}_k$, $\mathbf{P}_j$, $\mathbf{P}_k$, $\mathbf{c}_j^w$, $\mathbf{c}_k^w$, $\widehat{\mathbf{f}}_j^w$, $\widehat{\mathbf{f}}_k^w$, $\delta_\mathrm{epipolar}$, $\delta_\mathrm{lower}$, $\delta_\mathrm{upper}$, $\delta_{2D}$.}
\KwOutput{$\mathbf{x}_\mathrm{mid}^w$, $\widehat{\mathbf{f}}_j^w$, $\widehat{\mathbf{f}}_k^w$, $b_\mathrm{good}$.}

$b_\mathrm{good}\gets \text{false}$;\quad $\mathbf{x}_\mathrm{mid}^w\gets\mathbf{0}$;

\lIf {$\widehat{\mathbf{f}}_j^w=\mathbf{0}$}
{compute $\widehat{\mathbf{f}}^w_j$ using \eqref{eq:world_ray};}

\lIf {$\widehat{\mathbf{f}}_k^w=\mathbf{0}$}
{compute $\widehat{\mathbf{f}}^w_k$ using \eqref{eq:world_ray};}

$\mathbf{t}_{jk}^w\gets \mathbf{c}_j^w-\mathbf{c}_k^w$; \ \ 
$\widehat{\mathbf{t}}_{jk}^w\gets\mathbf{t}_{jk}^w/\left\Vert\mathbf{t}_{jk}^w\right\Vert$;

\tcc{Check the normalized epipolar error.}
  
compute $e_{jk}$ using \eqref{eq:epipolar_error};
  
\lIf {$e_{jk}>\delta_\mathrm{epipolar}$}{go to Line \ref{line:midpoint_return};} 

\tcc{Check the parallax angle.}

compute $p_{jk}$ using \eqref{eq:p};

\lIf {$p_{jk}<\delta_\mathrm{lower} \lor p_{jk} >\delta_\mathrm{upper}$}{go to Line \ref{line:midpoint_return};}

\tcc{Check additional degeneracy.}

compute $q_{jk}$ and $r_{jk}$ using \eqref{eq:qr};

\lIf {$|q_{jk}|>\delta_\mathrm{upper} \lor |r_{jk}| >\delta_\mathrm{upper}$}{go to Line \ref{line:midpoint_return};\hspace{-5em} } 

\tcc{Check the signs of anchor depths.}

compute $\mu_j$ and $\mu_k$ using \eqref{eq:mu};

\lIf {$\mu_j<0 \lor \mu_k<0$}{go to Line \ref{line:midpoint_return};} 

\tcc{Compute the midpoint.}

compute $s_{jk}$, $\lambda_j$, $\lambda_k$ and $\mathbf{x}_\mathrm{mid}^w$ using \eqref{eq:s}, \eqref{eq:lambdas}, \eqref{eq:x_mid};

\tcc{Check the cheirality.}

$\mathbf{x}_j\gets\mathbf{P}_j\begin{bmatrix}\mathbf{x}_\mathrm{mid}^w \\ 1 \end{bmatrix}$; \ \ 
$\mathbf{x}_k\gets\mathbf{P}_k\begin{bmatrix}\mathbf{x}_\mathrm{mid}^w \\ 1 \end{bmatrix}$;  

\lIf{$\left(\mathbf{x}_j\right)_3<0\lor \left(\mathbf{x}_k\right)_3<0$}{go to Line \ref{line:midpoint_return};}

\tcc{Check the reprojection error.}

$\mathbf{e}_j\gets\begin{bmatrix}\mathbf{u}_j\\1\end{bmatrix}-\displaystyle\frac{\mathbf{K}_j\mathbf{x}_j}{(\mathbf{x}_j)_3}$; \ \ 
$\mathbf{e}_k\gets\begin{bmatrix}\mathbf{u}_k\\1\end{bmatrix}-\displaystyle\frac{\mathbf{K}_k\mathbf{x}_k}{(\mathbf{x}_k)_3}$;

\lIf{$\mathbf{e}_j^\top\mathbf{e}_j>\delta^2_{2D} \lor \mathbf{e}_k^\top\mathbf{e}_k>\delta^2_{2D}$}{go to Line \ref{line:midpoint_return};}

$b_\mathrm{good}\gets\text{true}$;

\Return $\mathbf{x}_\mathrm{mid}^w$, $\widehat{\mathbf{f}}_i^w$, $\widehat{\mathbf{f}}_j^w$, $b_\mathrm{good}$; \label{line:midpoint_return}

\end{algorithm}

To obtain the initial solution and inlier set, we perform two-view RANSAC as in \cite{sfm_revisited}.
Our method has two notable differences to \cite{sfm_revisited}:
First, instead of the DLT method \cite{hartley_triangulation}, we use the midpoint method \cite{beardsley_midpoint2, hartley_triangulation}, which is faster and as accurate unless the parallax is very low \cite{hartley_triangulation, triangulation_why_optimize}.
Second, we prescreen the samples before performing the two-view triangulation.
Our method consists of the following steps:

\vspace{0.2em}
\textbf{1. Check the normalized epipolar error.}\\
Let camera $j$ and $k$ be the two-view sample. 
The normalized epipolar error \cite{geometric_interpretations_of_the_normalized_epipolar_error} of the sample is defined as
\vspace{-0.3em}
\begin{equation}
\label{eq:epipolar_error}
    e_{jk}:=\left|\widehat{\mathbf{t}}^w_{jk}\cdot\left(\widehat{\mathbf{f}}^w_j\times\widehat{\mathbf{f}}^w_k\right)\right|, 
\vspace{-0.7em}
\end{equation}
where $\widehat{\mathbf{t}}^w_{jk}$ is the unit vector of $\mathbf{t}_{jk}^w=\mathbf{c}_j^w-\mathbf{c}_k^w$ and $\widehat{\mathbf{f}}^w_j$, $\widehat{\mathbf{f}}^w_k$ are the corresponding rays in the world frame given by \eqref{eq:world_ray}.
If $\widehat{\mathbf{f}}^w_j$ and $\widehat{\mathbf{f}}^w_k$ are both inliers, then $e_{jk}$ must be small \cite{eight_point}.

\vspace{0.5em}
\textbf{2. Check the parallax angle.}\\
The raw parallax \cite{triangulation_why_optimize} defined as
$\beta_{jk}:=\angle(\widehat{\mathbf{f}}^w_j,  \widehat{\mathbf{f}}^w_k)$ is a rough estimate of the parallax angle if $\widehat{\mathbf{f}}^w_j$ and $\widehat{\mathbf{f}}^w_k$ are both inliers.
If $\beta_{jk}$ is too small, the triangulation is inaccurate \cite{triangulation_why_optimize}.
If it is too large, the sample most likely contains an outlier, because such a point is rarely matched in practice due to a severe viewpoint change. 
We check $\beta_{jk}$ from its cosine:
\vspace{-0.2em}
\begin{equation}
\label{eq:p}
    p_{jk}:=\widehat{\mathbf{f}}_j^w\cdot\widehat{\mathbf{f}}_k^w
\end{equation}

\textbf{3. Check additional degeneracy.}\\[0.2em]
Likewise, if $\angle(\widehat{\mathbf{f}}_j^w, \widehat{\mathbf{t}}_{jk}^w)$ or $\angle(\widehat{\mathbf{f}}_k^w, \widehat{\mathbf{t}}_{jk}^w)$ is too small, the epipolar geometry degenerates (see Fig. \ref{fig:midpoint}).
To avoid degeneracy, we also check these angles from their cosines:
\vspace{-0.3em}
\begin{equation}
    q_{jk}:=\widehat{\mathbf{f}}_j^w\cdot\widehat{\mathbf{t}}_{jk}^w, \ \  
    r_{jk}:=\widehat{\mathbf{f}}_k^w\cdot\widehat{\mathbf{t}}_{jk}^w. \label{eq:qr}
\end{equation}

\textbf{4. Check the depths of the midpoint anchors.}\\
Let $\lambda_j$ and $\lambda_k$ be the depths of the midpoint anchors (see Fig. \ref{fig:midpoint}).
In the supplementary material, we show that
\vspace{-0.5em}
\begin{equation}
\lambda_j=s_{jk}\mu_j, \quad 
\lambda_k=s_{jk}\mu_k. \label{eq:lambdas}
\vspace{-0.5em}
\end{equation}
where
\vspace{-1em}
\begin{gather}
    s_{jk}:=\left\Vert\mathbf{t}^w_{jk}\right\Vert/\left(1-p_{jk}^2\right), \label{eq:s}\\
    \mu_{j}:=p_{jk}r_{jk}-q_{jk}, \quad \mu_{k}:=-p_{jk}q_{jk}+r_{jk}. \label{eq:mu}
\vspace{-0.5em}
\end{gather}
Since $s_{jk} \geq 0$, we check the signs of $\mu_j$ and $\mu_k$ to ensure that $\lambda_j$ and $\lambda_k$ are both positive.

\textbf{5. Evaluate the midpoint w.r.t. the two views.}\\
Only when the two-view sample passes all of the aforementioned checks, we compute the midpoint:
\vspace{-0.2em}
\begin{equation}
\label{eq:x_mid}
    \mathbf{x}^w_\mathrm{mid}=0.5\left(\mathbf{c}_j^w+\lambda_j\widehat{\mathbf{f}}_j^w+\mathbf{c}_k^w+\lambda_k\widehat{\mathbf{f}}_k^w\right).
\vspace{-0.2em}
\end{equation}
Then, we check the cheirality and reprojection errors in the two views.
The entire procedure is detailed in Alg. \ref{al:midpoint}.

\begin{fontspace}{0.99}{1}
Each midpoint from the two-view samples becomes a hypothesis for $\mathbf{x}^w$ and is scored based on its reprojection error and cheirality. 
Specifically, we use the approach of \cite{msac} and find the hypothesis that minimizes the following cost:
\end{fontspace}
\vspace{-1em}
\begin{equation}
\label{eq:cost}
    C=\sum_{i=1}^n r_i^2 \quad \text{with} \quad
    r_i = 
    \begin{cases}
    \left(\mathbf{e}_{2D}\right)_i & \text{ if } i\in\mathcal{I},\\
    \delta_{2D} & \text{ otherwise.}
    \end{cases}
\end{equation}
Once we find a hypothesis with smaller cost, we update the inlier ratio based on its support set $\mathcal{I}$ and recompute the required number of samples to be drawn (adaptive stopping criterion \cite{torr_1998_robust, usac, sfm_revisited}).
This is done in Line \ref{line:adaptive_stopping} of Alg. \ref{al:proposed}.

%%%% Revert margins to default
\setlength{\textfloatsep}{\textfloatsepsave}

\subsection{Iterative Local Optimization}
\label{subsec:local_optimization}
Once we have the initial triangulation result and the inlier set from the two-view RANSAC, we perform local optimization for refinement.
Our approach is similar to \cite{loransac}, except that we perform the optimization only at the end of RANSAC.
We compare three optimization methods:
\noindent
\begin{itemize}
    \item \texttt{DLT} and \texttt{LinLS} \cite{hartley_triangulation}:
    These two linear methods minimize the algebraic errors in closed form.
    For the formal descriptions, we refer to \cite{hartley_triangulation}.
    They were originally developed for two-view triangulation, but they can be easily extended to multiple views.
    We construct the linear system with only the inliers, solve it and update the inlier set.
    This is repeated until the inlier set converges.
    \item \texttt{GN}: This nonlinear method minimizes the geometric errors using the Gauss-Newton algorithm.
    After each update of the solution, we update the inlier set.
\end{itemize}

The \texttt{GN} method requires the computation of the Jacobian matrix $\mathbf{J}$ in each iteration.
In the following, we present an efficient method for computing $\mathbf{J}$.
Recall that we are minimizing $\mathbf{e}_{2D}^\top\mathbf{e}_{2D}$, which we know from \eqref{eq:reproj_error} is equal to $\mathbf{r}^\top\mathbf{r}$, where $\mathbf{r}=\left[(u_\mathrm{error})_1, (v_\mathrm{error})_1, \cdots, (u_\mathrm{error})_n, (v_\mathrm{error})_n\right]^\top$.
This means that we can obtain $\mathbf{J}$ by stacking 
\begin{equation}
\label{eq:Jacobian}
    \mathbf{J}_i=
    \begin{bmatrix}
        \displaystyle\frac{\partial(u_\mathrm{error})_i}{\partial x^w_\mathrm{est}} 
        &
        \displaystyle\frac{\partial(u_\mathrm{error})_i}{\partial y^w_\mathrm{est}}
        &
        \displaystyle\frac{\partial(u_\mathrm{error})_i}{\partial z^w_\mathrm{est}}\\
        \displaystyle\frac{\partial(v_\mathrm{error})_i}{\partial x^w_\mathrm{est}} 
        &
        \displaystyle\frac{\partial(v_\mathrm{error})_i}{\partial y^w_\mathrm{est}}
        &
        \displaystyle\frac{\partial(v_\mathrm{error})_i}{\partial z^w_\mathrm{est}}
    \end{bmatrix}
\end{equation}
for all $i\in\mathcal{I}$.
We now define the following vectors:
\begin{flalign}
    &\mathbf{b}_{1i} := r_{i11}\left[0,r_{i32},r_{i33},t_{i3}\right]^\top\hspace{-0.5em}-r_{i31}\left[0,r_{i12},r_{i13},t_{i1}\right]^\top\hspace{-0.5em},\hspace{-5em}& \label{eq:b1}\\
    &\mathbf{b}_{2i} := r_{i21}\left[0,r_{i32},r_{i33},t_{i3}\right]^\top\hspace{-0.5em}-r_{i31}\left[0,r_{i22},r_{i23},t_{i2}\right]^\top\hspace{-0.5em},\hspace{-5em}&\\
    &\mathbf{b}_{3i} := r_{i12}\left[r_{i31},0,r_{i33},t_{i3}\right]^\top\hspace{-0.5em}-r_{i32}\left[r_{i11},0,r_{i13},t_{i1}\right]^\top\hspace{-0.5em},\hspace{-5em}&\\
    &\mathbf{b}_{4i} := r_{i22}\left[r_{i31},0,r_{i33},t_{i3}\right]^\top\hspace{-0.5em}-r_{i32}\left[r_{i21},0,r_{i23},t_{i2}\right]^\top\hspace{-0.5em},\hspace{-5em}&\\
    &\mathbf{b}_{5i} := r_{i13}\left[r_{i31},r_{i32},0,t_{i3}\right]^\top\hspace{-0.5em}-r_{i33}\left[r_{i11},r_{i12},0,t_{i1}\right]^\top\hspace{-0.5em},\hspace{-5em}&\\
    &\mathbf{b}_{6i} := r_{i23}\left[r_{i31},r_{i32},0,t_{i3}\right]^\top\hspace{-0.5em}-r_{i33}\left[r_{i21},r_{i22},0,t_{i2}\right]^\top\hspace{-0.5em},\hspace{-5em}& \label{eq:b6}\\
    &\mathbf{a}_{1i}:=k_{i11}\mathbf{b}_{1i}+k_{i12}\mathbf{b}_{2i}, \ \ 
    \mathbf{a}_{2i}:=k_{i21}\mathbf{b}_{1i}+k_{i22}\mathbf{b}_{2i},\hspace{-5em}& \label{eq:a1}\\
    &\mathbf{a}_{3i}:=k_{i11}\mathbf{b}_{3i}+k_{i12}\mathbf{b}_{4i}, \ \ 
    \mathbf{a}_{4i}:=k_{i21}\mathbf{b}_{3i}+k_{i22}\mathbf{b}_{4i},\hspace{-5em}&\\
    &\mathbf{a}_{5i}:=k_{i11}\mathbf{b}_{5i}+k_{i12}\mathbf{b}_{6i}, \ \ 
    \mathbf{a}_{6i}:=k_{i21}\mathbf{b}_{5i}+k_{i22}\mathbf{b}_{6i},\hspace{-5em}& \label{eq:a6}
\end{flalign}
where $r_{ijk}$ and $k_{ijk}$ respectively indicate the elements of $\mathbf{R}_i$ and $\mathbf{K}_i$ at the $j$-th row and $k$-th column, and $t_{ij}$ indicate the $j$-th element of $\mathbf{t}_i$.
Then, we can rewrite \eqref{eq:Jacobian} as
\begin{gather}
    \mathbf{J}_i=\left((\mathbf{P}_i)_\mathrm{row3}\widetilde{\mathbf{x}}^w_\mathrm{est}\right)^{-2}\mathrm{vec}^{-1}_{2\times3}\left(
    \mathbf{A}_i^\top\widetilde{\mathbf{x}}^w_\mathrm{est}\right) \label{eq:Jacobian_fast}\\
    \text{with} \quad
    \mathbf{A}_i = 
    \left[
        \mathbf{a}_{1i} \ \mathbf{a}_{2i} \ \mathbf{a}_{3i} \ \mathbf{a}_{4i} \ \mathbf{a}_{5i} \ \mathbf{a}_{6i}
    \right].\label{eq:A}
\end{gather}
We provide the derivation in the supplementary material.
Since $\mathbf{P}_i$ and $\mathbf{A}_i$ can be precomputed independently of the point, the Jacobian can computed more efficiently.
Alg. \ref{al:gn} summarizes the \texttt{GN} method.

\begin{algorithm}[t]
\setstretch{0.95}
\caption{\texttt{GN} with an iterative update of the $\text{inlier set}\hspace{-10em}$}
\label{al:gn}
\DontPrintSemicolon
\KwInput{$\mathbf{x}^w_\mathrm{est}$, $\mathcal{V}$, $\mathcal{I}$, $\mathbf{M}_{1,2,\cdots,8}$, $\delta_{2D}$, $\delta_\mathrm{update}$, $\mathbf{P}_i$ and $\mathbf{A}_i$ for all $i\in\mathcal{V}$.}
\KwOutput{$\mathbf{x}^w_\mathrm{est}$, $\mathcal{I}$, $\overline{e}_{2D}$.}

$n_\mathrm{it}\gets0$; \ \
$\overline{e}_{2D}\gets0$; 

\While{$n_\mathrm{it}<10$}
{
    $n_\mathrm{it}\gets n_\mathrm{it}+1$;
    \ \
    $(\overline{e}_{2D})_\mathrm{prev} \gets \overline{e}_{2D}$;
    \ \
    $\mathcal{I}_\mathrm{prev}\gets\mathcal{I}$;
    
    \tcc{Obtain the residuals and Jacobian.\hspace{-5em}}
    
    obtain $\mathbf{r}$ by stacking $\begin{bmatrix}(\mathbf{M}_7)_i \\  (\mathbf{M}_8)_i\end{bmatrix}$ for all $i\in\mathcal{I}$;
    
    compute $\mathbf{J}_i$ using \eqref{eq:Jacobian_fast} for all $i\in\mathcal{I}$;
    
    obtain $\mathbf{J}$ by stacking $\mathbf{J}_i$ for all $i\in\mathcal{I}$;
    
    \tcc{Update the solution.}
    
    $\mathbf{x}^w_\mathrm{est}\gets\mathbf{x}^w_\mathrm{est}-\mathbf{J}^+\mathbf{r}$;
    
    \tcc{Update the inlier set.}
    
    compute $\mathbf{M}_6$, $\mathbf{M}_7$, $\mathbf{M}_8$ using \eqref{eq:M6}--\eqref{eq:M8};
    
    compute $\mathbf{e}_{2D}$ and $\mathcal{I}$ using \eqref{eq:reproj_error_fast} and \eqref{eq:inlier_set};
        
    \tcc{Check the convergence.}
    
    $\overline{e}_{2D}\gets \frac{1}{|\mathcal{I}|}\sum_{i\in\mathcal{I}} \left(\mathbf{e}_{2D}\right)_i$;
    
    \If{$\mathcal{I}=\mathcal{I}_\mathrm{prev} \land |\overline{e}_{2D}-(\overline{e}_{2D})_\mathrm{prev}|<\delta_\mathrm{update}$}{break;}
}
\Return $\mathbf{x}_\mathrm{est}^w$, $\mathcal{I}$, $\overline{e}_{2D}$; \label{line:gn_return}
\end{algorithm}
\begin{algorithm}[t]
\setstretch{0.95}
\caption{Proposed 3D uncertainty estimation}
\label{al:uncertainty}
\DontPrintSemicolon
\KwInput{$G$, $\mathcal{I}$, $\overline{e}_{2D}$, $\delta_\mathrm{pair}$,  $\mathbf{u}_i$, $\widehat{\mathbf{f}}_i^w$, $\mathbf{K}_i$, $\mathbf{R}_i$ for all $i\in\mathcal{V}.\hspace{-5em}$}
\KwOutput{$\sigma_{3D}$.}

\tcc{Estimate the maximum parallax angle.}

$p_\mathrm{min}\gets \infty$, \ \  $\delta_\mathrm{pair}\gets\min\left(\delta_\mathrm{pair}, |\mathcal{I}|(|\mathcal{I}|-1)/2\right)$; \ \ 
$n_\mathrm{pair}\gets 0$

\While{$n_\mathrm{pair}<\delta_\mathrm{pair}$}
{
    $n_\mathrm{pair}\gets n_\mathrm{pair}+1$;
    
    Pick a random pair of views $j,k\in\mathcal{I}$;
    
    \lIf {$\widehat{\mathbf{f}}_j^w=\mathbf{0}$}
    {compute $\widehat{\mathbf{f}}^w_j$ using \eqref{eq:world_ray};}
    
    \lIf {$\widehat{\mathbf{f}}_k^w=\mathbf{0}$}
    {compute $\widehat{\mathbf{f}}^w_k$ using \eqref{eq:world_ray};}

    $p\gets\left|\widehat{\mathbf{f}}_j^w\cdot\widehat{\mathbf{f}}_k^w\right|$;
    
    \lIf{$p<p_\mathrm{min}$}{$p_\mathrm{min}\gets p$;}
}

$\beta_\mathrm{max}\gets \cos^{-1}(p_\mathrm{min})$;

\tcc{Interpolate the uncertainty.}

$n_\mathrm{in}\gets\min(|\mathcal{I}|, 50)$; \ \ 
$\overline{e}_{2D}\gets\min(\overline{e}_{2D}, 20\text{ pix})$; \ \ 
$\beta_\mathrm{max}\gets\min(\beta_\mathrm{max}, 20^\circ)$;

Obtain $\sigma_{3D}$ by performing trilinear interpolation on the 3D grid $G$ at $(n_\mathrm{in}, \overline{e}_{2D}, \beta_\mathrm{max})$;

\Return $\sigma_{3D}$; 
\end{algorithm}
\begin{figure*}[t]
\vspace{-1em}
 \centering
 \includegraphics[width=\textwidth]{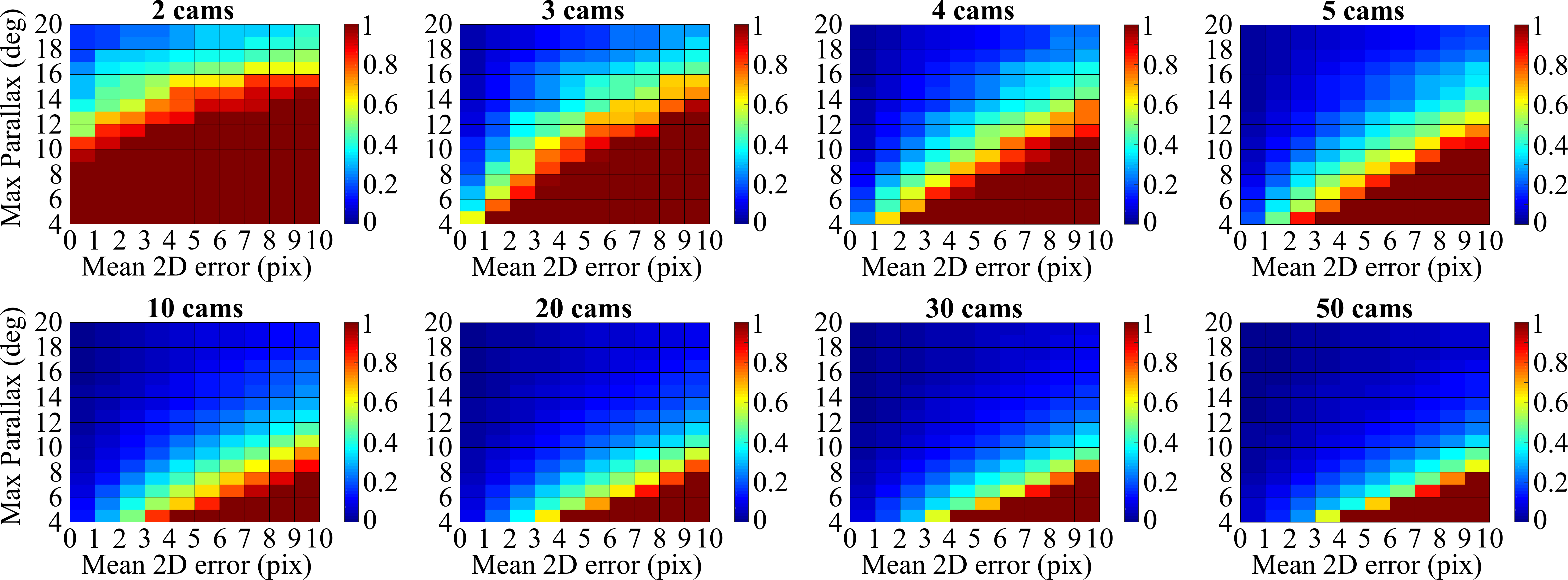}
\caption{RMS of the 3D errors for different numbers of cameras, maximum parallax angles and mean 2D errors.
We only present the smoothed results up to 10 pixel error for the selected numbers of cameras (see the supplementary material for the full results).
Any cell value above one unit is considered highly inaccurate, and thus truncated.
One unit corresponds to the geometric span of the cameras.
 }
\label{fig:uncertainty}
\end{figure*}

\subsection{Practical Uncertainty Estimation}
\label{subsec:uncertainty}
\vspace{-0.3em}
We model the uncertainty of the triangulated point $\mathbf{x}^w_\mathrm{est}$ as a function of three factors: the number of inlying views ($|\mathcal{I}|$), the mean reprojection error in those views ($\overline{e}_\text{2D}$) and the maximum parallax angle ($\beta_\text{max}$) defined by \eqref{eq:beta_max}.

To this end, we run a large number of simulations in various settings and aggregate the 3D errors for each different range of factors (see Fig. \ref{fig:uncertainty}).
We then store these data on a 3D regular grid $G$ that maps $(|\mathcal{I}|, \overline{e}_\text{2D}, \beta_\text{max})$ to the uncertainty $\sigma_{3D}$.
At test time, we estimate the uncertainty by performing trilinear interpolation on this grid.

We point out two things in our implementation:
First, to reduce the small sample bias in $G$, we perform monotone smoothing that enforces $\sigma_{3D}$ to increase with $\overline{e}_\text{2D}$ and decreases with $|\mathcal{I}|$ and $\beta_\text{max}$.
The smoothing method is described and demonstrated in the supplementary material.
Second, we limit the number of pairs we evaluate for computing $\beta_\mathrm{max}$ in \eqref{eq:beta_max2}.
This curbs the computational cost when $\mathcal{I}$ is very large.
Alg. \ref{al:uncertainty} summarizes the procedure.

\section{Results}
\subsection{Uncertainty Estimation}
\label{subsec:result_uncertainty}
To find out how the different factors impact the 3D accuracy of triangulation, we run a large number of simulations in various settings configured by the following parameters:
\begin{itemize}
    \item $n$: number of cameras observing the point.
    \item $d$: distance between the ground-truth point and the origin.
    \item $\sigma$: std. dev. of Gaussian noise in the image coordinates.
    \item $n_\text{run}$: number of independent simulation runs for each configuration $(n,d,\sigma)$.
\end{itemize}
The parameter values are specified in the supplementary material.
The simulations are generated as follows:
We create $n$ cameras, $n-2$ of those randomly located inside a sphere of unit diameter at the origin.
We place one of the two remaining cameras at a random point on the sphere's surface and the other at its antipode.
This ensures that the geometric span of the cameras is equal to one unit.
The size and the focal length of the images are set to $640\times480$ and $525$ pixel, respectively, the same as those of \cite{tum_rgbd_dataset}.
Next, we create a point at $[0,0,d]^\top$ and orient the cameras randomly until the point is visible in all images. 
Then, we add the image noise of $\mathcal{N}(0,\sigma^2)$ to perturb the image coordinates.

For triangulation, we initialize the point using the \texttt{DLT} method and refine it using the \texttt{GN} method.
In this experiment, we assume that all points are always inliers, so we do not update the inlier set during the optimization.
Fig. \ref{fig:uncertainty} shows the 3D error distribution with respect to different numbers of cameras ($n$), mean reprojection errors ($\overline{e}_\text{2D}$) and maximum parallax angle ($\beta_\text{max}$).
In general, we observe that the 3D accuracy improves with more cameras, smaller $\overline{e}_\text{2D}$ and larger $\beta_\text{max}$.
However, this effect diminishes past a certain level.
For example, the difference between the 30 and 50 cameras is much smaller than the difference between the 2 and 3.
Also, when $\beta_\text{max}$ is sufficiently large, the 3D accuracy is less sensitive to the change in $n$, $\overline{e}_\text{2D}$ and $\beta_\text{max}$.

Fig. \ref{fig:uncertainty} clearly indicates that we must take into account all these three factors when estimating the 3D uncertainty of a triangulated point.
Marginalizing any one of them would reduce the accuracy.
This observation agrees with our intuition, as each factor conveys important independent information about the given triangulation problem.
%In the next section, we will show that the estimated uncertainty can be used for controlling the 3D accuracy. 

\begin{figure*}[t]
 \centering
 \includegraphics[width=\textwidth]{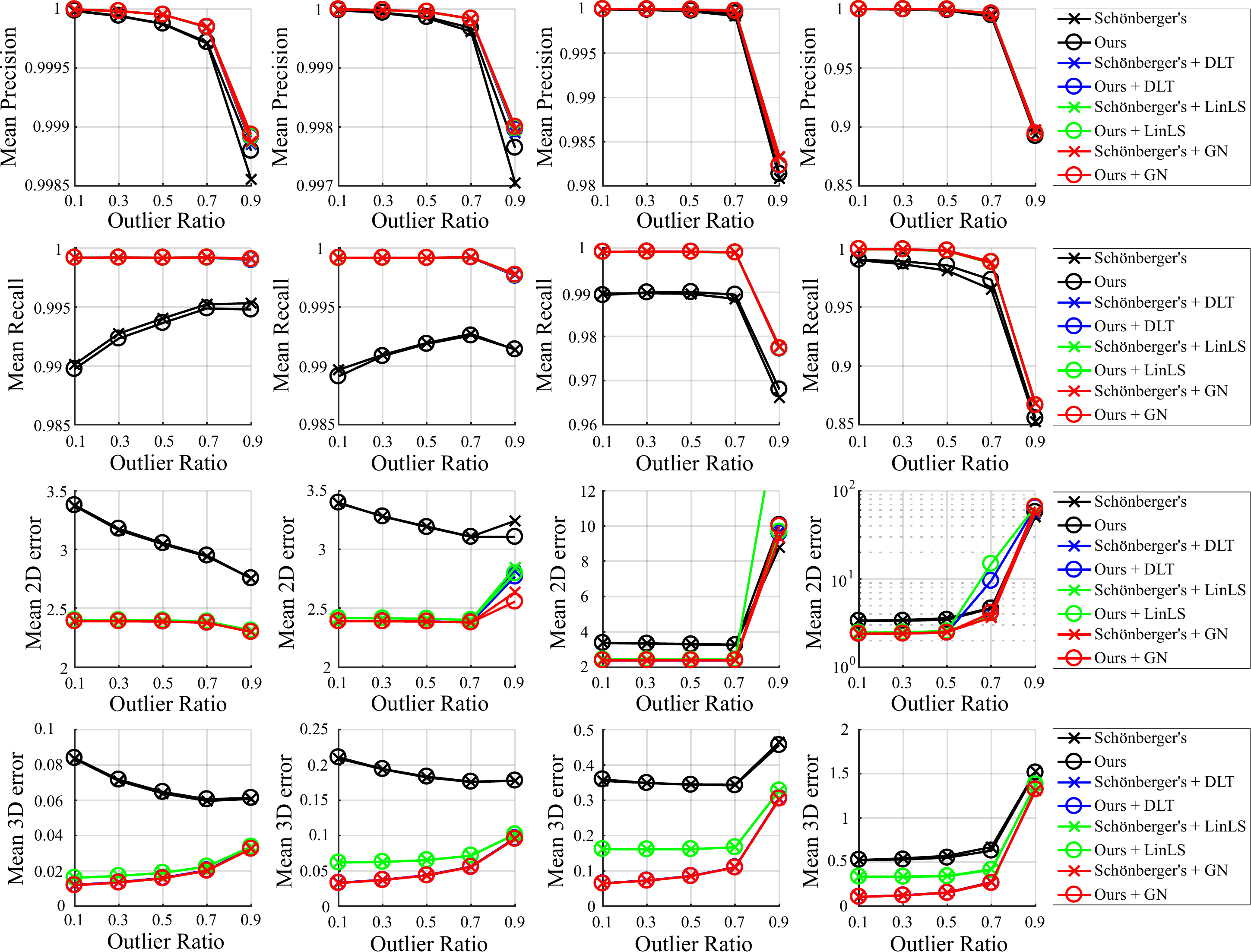}
\caption{Triangulation Performance.
From left to right, the columns correspond to the different point distances (3, 5, 7, 9 unit).
One unit corresponds to the geometric span of the cameras.
%In the last row, we plot the percentage of the points excluded from the result, either due to the final pruning or the lack of inliers from the RANSAC procedure.
The mean 2D error is computed with respect to \textbf{all true inlying} observations.
 }
\label{fig:performance}
\end{figure*}

\subsection{Triangulation Performance}
\label{subsec:triangulation_performance}
We evaluate the performance of our method on synthetic data.
The simulation is configured in a similar way as in the previous section.
The difference is that we set $n=100$, $d=\{3,5,7,9\}$, $\sigma=3$ pixel, $n_\text{run}=100$ thousand, and we perturb some of the measurements by more than 10 pixel, turning them into outliers.
The outlier ratio is set to $10$, $30$, $50$, $70$ and $90$ percent. 
Varying $d$ and the outlier ratio results in $4
\times5$ configurations, so in total, this amounts to two million unique triangulation problems.

On this dataset, we compare our method against the state of the art (\cite{sfm_revisited} by Sch\"{o}nberger and Frahm) with and without the local optimization (\texttt{DLT}, \texttt{LinLS} and \texttt{GN}).
In Alg. \ref{al:proposed}, we set $\eta{\hspace{0.1em}=\hspace{0.1em}}0.99$, $\delta_{2D}{\hspace{0.1em}=\hspace{0.1em}}10$ pix, $\delta_\mathrm{epipolar}{\hspace{0.1em}=\hspace{0.1em}}0.01$, $\delta_\mathrm{update}{\hspace{0.1em}=\hspace{0.1em}}0.1$ pix, $\delta_\mathrm{lower}{\hspace{0.1em}=\hspace{0.1em}}0$,  $\delta_\mathrm{upper}{\hspace{0.1em}=\hspace{0.1em}}\cos(4^\circ)$ and $\delta_\mathrm{pair}{\hspace{0.1em}=\hspace{0.1em}}100$.
Fig. \ref{fig:performance} shows the results.
On average, ours and \cite{sfm_revisited} perform similarly (but ours is faster, as will be shown later).
For both methods, the local optimization substantially improves the 2D and 3D accuracy.
Thanks to the iterative update of the inlier set, we also see a significant gain in recall. 
Among the optimization methods, \texttt{DLT} and \texttt{GN} show similar performance in all criteria, while \texttt{LinLS} exhibits larger 3D error than the other two.
We provide a closer comparison between \texttt{DLT} and \texttt{GN} in the next section.

In general, when the point is far and the outlier ratio is high, the performance degrades for all methods.
At any fixed outlier ratio, we observe that the 3D error tends to grow with the point distance.
%This agrees with our finding in the previous section where the 3D error was shown to be greater at lower parallax.
However, the same cannot be said for the 2D error.
This is because given sufficient parallax, the 2D accuracy is mostly influenced by the image noise statistics, rather than the geometric configurations.

\begin{figure}[t]
 \centering
 \includegraphics[width=0.49\textwidth]{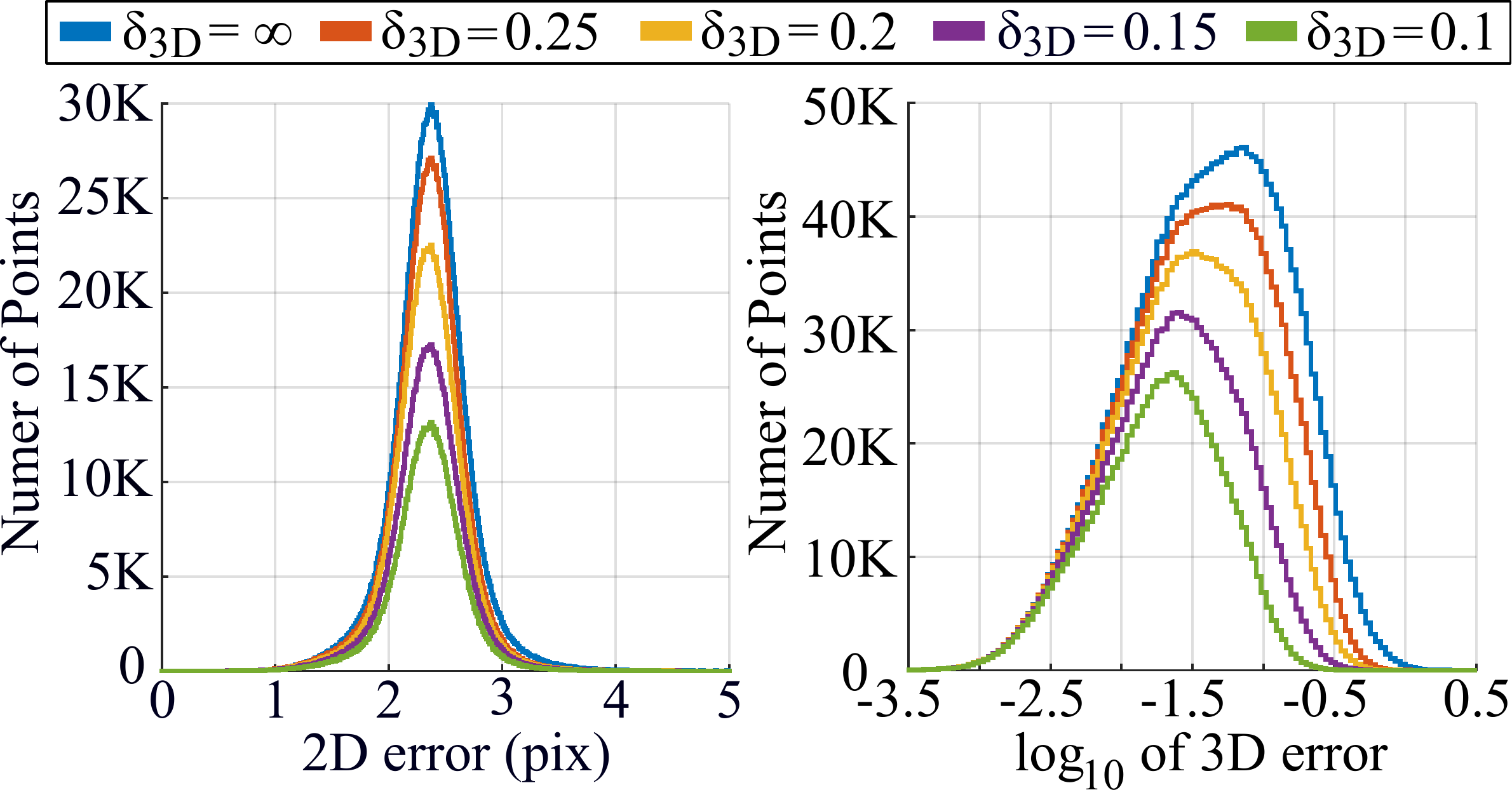}
\caption{Error histograms of the triangulated points with the mean 2D error $<$ 5 pix and the estimated uncertainty $\sigma_{3D}<\delta_{3D}$.
 }
\label{fig:prune2}
\end{figure}

\begin{figure}[t]
 \centering
 \includegraphics[width=0.4\textwidth]{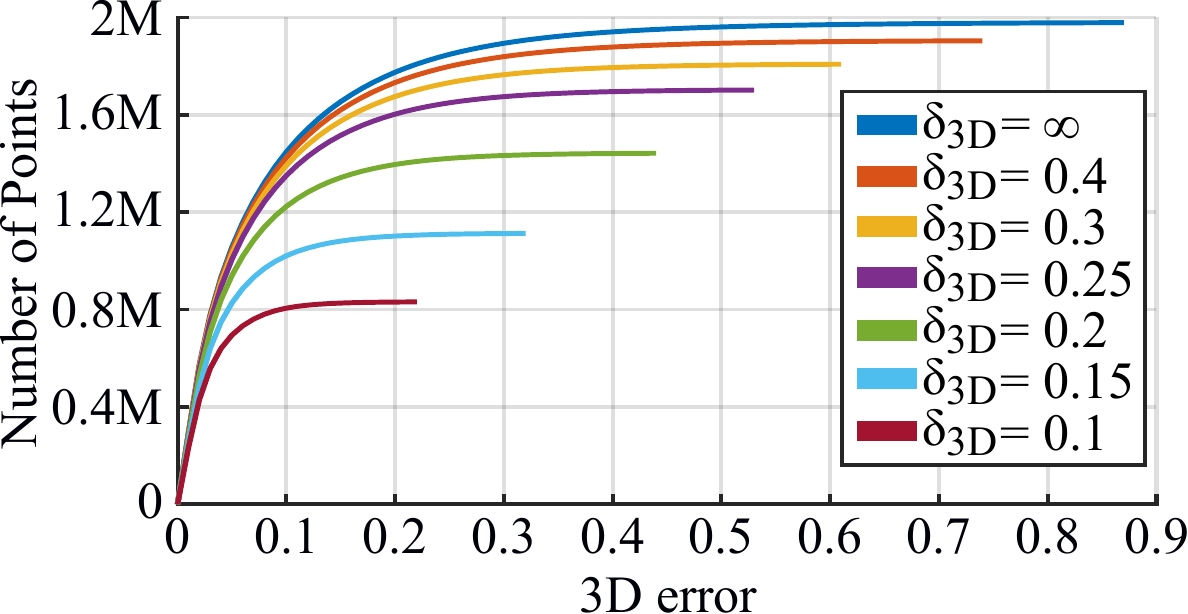}
\caption{Cumulative error plots of the triangulated points with the mean 2D error $<$ 5 pix and the estimated uncertainty $\sigma_{3D}<\delta_{3D}$.
We truncate each curve at 99.9\% accumulation.
 }
\label{fig:prune3}
\end{figure}

\begin{figure}[t]
 \centering
 \includegraphics[width=0.4\textwidth]{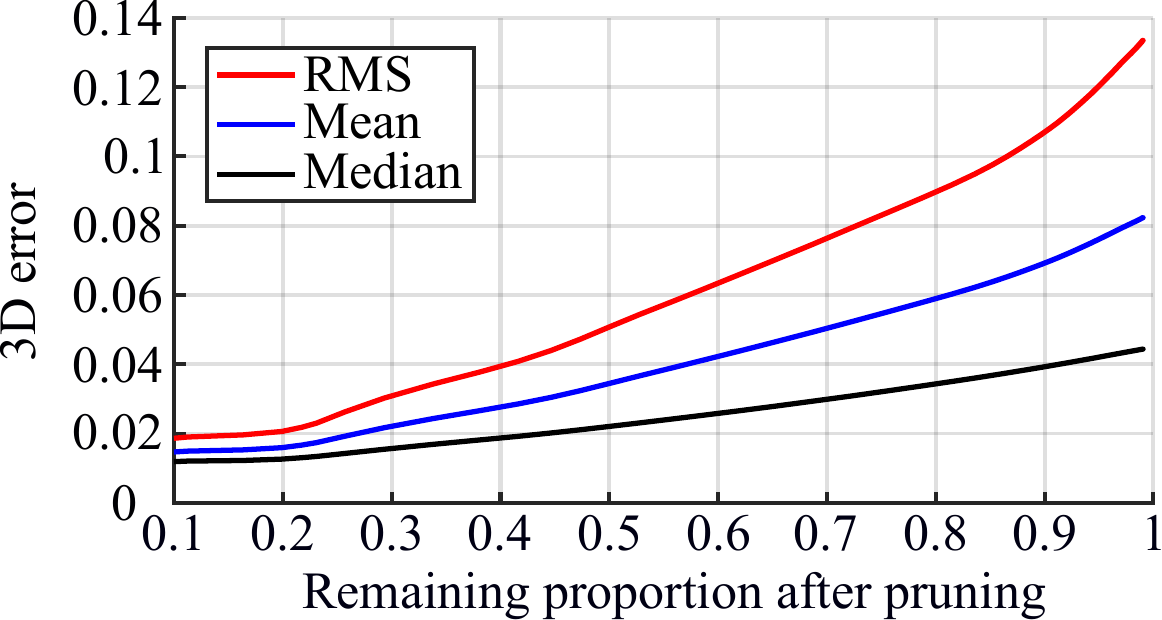}
\caption{Trade-off between the 3D error and the number of points by varying the uncertainty threshold for pruning. 
As in Fig. \ref{fig:prune2} and \ref{fig:prune3}, we only consider the points with the mean 2D error $<$ 5 pix.
 }
\label{fig:prune4}
\end{figure}

We also evaluate the accuracy after pruning the most uncertain points using our method (Sect. \ref{subsec:result_uncertainty}).
In Fig. \ref{fig:prune2}, we plot the error histograms of the points with different levels of the estimated 3D uncertainty ($\sigma_{3D}$).
It shows that with a smaller threshold on $\sigma_{3D}$, we get to prune more points with larger 3D error.
Fig. \ref{fig:prune3} shows the cumulative 3D error plots.
It illustrates that thresholding on $\sigma_{3D}$ gives us some control over the upper bound of the 3D error.
As a result, we are able to trade off the number points for 3D accuracy by varying the threshold level.
This is shown in Fig. \ref{fig:prune4}.


To compare the timings, all methods are implemented in MATLAB and run on a laptop CPU (Intel i7-4810MQ, 2.8GHz).
Tab. \ref{tab:timings} provides the relative speed of our two-view RANSAC compared to \cite{sfm_revisited}.
It shows that ours is faster, especially when the point is far and the outlier ratio is high.
This demonstrates the advantage of the early termination of two-view triangulation (Alg. \ref{al:midpoint}).
In Tab. \ref{tab:timings2}, we present the timings of the local optimization and uncertainty estimation.
We found that \texttt{DLT} is slightly faster than \texttt{LinLS} and almost twice faster than \texttt{GN}.

\begin{table}[t]
\small
\begin{center}
\setlength\tabcolsep{5pt}
\begin{tabular}{c|c|c|c|c|}
\cline{2-5}
& $d=3$ & $d=5$ & $d=7$ & $d=9$ \\
\hline
\multicolumn{1}{|c|}{\multirow{2}{*}{$\text{OR}=10\%$}} & 4.15, 4.15 & 4.43, 4.32 & 4.81, 4.40 & 5.90, 4.47 \\
\multicolumn{1}{|c|}{} & $(\bm{\times1.00})$& $(\bm{\times1.03})$ & $(\bm{\times1.09})$ &$(\bm{\times1.32})$ \\
\hline
\multicolumn{1}{|c|}{\multirow{2}{*}{$\text{OR}=30\%$}}& 4.61, 4.39& 4.80, 4.45 & 4.78, 4.08 & 7.33, 4.79 \\
\multicolumn{1}{|c|}{} & $(\bm{\times1.05})$& $(\bm{\times1.08})$ & $(\bm{\times1.17})$ &$(\bm{\times1.53})$ \\
\hline
\multicolumn{1}{|c|}{\multirow{2}{*}{$\text{OR}=50\%$}}& 5.28, 4.56 & 5.64, 4.72 & 5.87, 4.38 & 10.4, 5.22 \\
\multicolumn{1}{|c|}{} & $(\bm{\times1.16})$& $(\bm{\times1.20})$ & $(\bm{\times1.34})$ &$(\bm{\times1.99})$ \\
\hline
\multicolumn{1}{|c|}{\multirow{2}{*}{$\text{OR}=70\%$}}& 7.46, 5.09& 8.06, 5.30 & 9.13, 4.93 & 19.9, 6.45 \\
\multicolumn{1}{|c|}{} & $(\bm{\times1.46})$& $(\bm{\times1.52})$ & $(\bm{\times1.85})$ &$(\bm{\times3.09})$ \\
\hline
\multicolumn{1}{|c|}{\multirow{2}{*}{$\text{OR}=90\%$}}& 28.6, 7.64& 31.8, 8.14 & 40.8, 8.86 & 72.6, 13.1 \\
\multicolumn{1}{|c|}{} & $(\bm{\times3.74})$& $(\bm{\times3.91})$ & $(\bm{\times4.61})$ &$(\bm{\times5.56})$ \\
\hline
\end{tabular}
\end{center}
\caption{
RANSAC time per point (ms). 
The two entries respectively correspond to \cite{sfm_revisited} and ours without local optimization.
The relative speed of ours compared to \cite{sfm_revisited} is given in parentheses.}
\label{tab:timings}
\end{table}
\begin{table}[ht]
\small
\begin{center}
\begin{tabular}{c|c|c|c|c|}
\cline{2-5}
& DLT & LinLS & GN & Uncertainty Est. \\
\hline
\multicolumn{1}{|c|}{$\text{OR}=10\%$} & \textbf{1.57} & 1.62 & 3.06 & 1.64\\
\hline
\multicolumn{1}{|c|}{$\text{OR}=30\%$}& \textbf{1.18} & 1.27 & 2.39 & 1.44\\
\hline
\multicolumn{1}{|c|}{$\text{OR}=50\%$}& \textbf{0.84} & 0.95 & 1.78 & 1.25\\
\hline
\multicolumn{1}{|c|}{$\text{OR}=70\%$}& \textbf{0.57} & 0.64 & 1.17 & 1.04\\
\hline
\multicolumn{1}{|c|}{$\text{OR}=90\%$}& \textbf{0.25} & 0.29 & 0.50 & 0.42\\
\hline
\end{tabular}
\end{center}
\caption{
Optimization and uncertainty estimation time per point (ms). 
The fastest optimization result is shown in bold.
}
\label{tab:timings2}
\end{table}
\begin{figure}[t]
 \centering
 \includegraphics[width=0.48\textwidth]{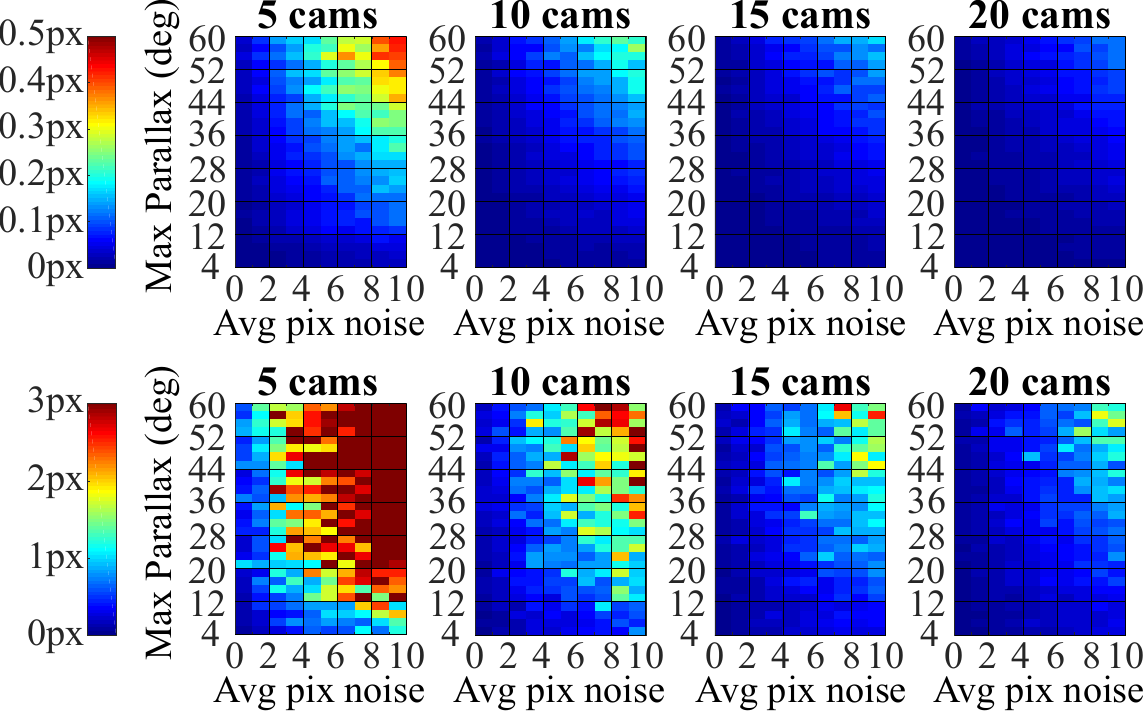}
\caption{\textbf{Top row:} Mean decrease of 2D error in $L_1$ norm by performing \texttt{GN} in addition to \texttt{DLT}, \ie, $\overline{e}_\text{2D DLT}-\overline{e}_\text{2D GN}$, for different configurations (see the supplementary material for the full results). 
The redder the color, the more accurate \texttt{GN} is than \texttt{DLT}.
\textbf{Bottom row:} Maximum decrease of 2D error, \ie, $\max(e_\text{2D DLT}-e_\text{2D GN})$.
}
\label{fig:dlt_vs_gn}
\end{figure}

\subsection{DLT vs. Gauss-Newton Method}
To compare the accuracy of \texttt{DLT} and \texttt{GN} more closely, we perform additional simulations in outlier-free scenarios.
The simulation is set up in a similar way as in Section \ref{subsec:result_uncertainty} (see the supplementary material for details).

In terms of 3D accuracy, we found that the two methods perform almost equally most of the time.
The comparison is inconsistent only when the maximum parallax angle is very small (less than 6 deg or so).
%This is because the geometric configuration degenerates at very low parallax, and the triangulation problem becomes ill-conditioned.
We show this result in the supplementary material.

As for the 2D accuracy, the difference is sometimes noticeable.
Fig. \ref{fig:dlt_vs_gn} shows the mean and the maximum difference of 2D error.
On average, \texttt{GN} offers less gain for more cameras, smaller noise and lower parallax. 
This explains why we could not see the difference between \texttt{DLT} and \texttt{GN} in Fig. \ref{fig:performance}.
However, the bottom row of Fig. \ref{fig:dlt_vs_gn} reveals that \texttt{GN} sometimes provides a significant gain over \texttt{DLT} even when the average difference is small.

\begin{figure*}[t]
 \centering
 \includegraphics[width=0.97\textwidth]{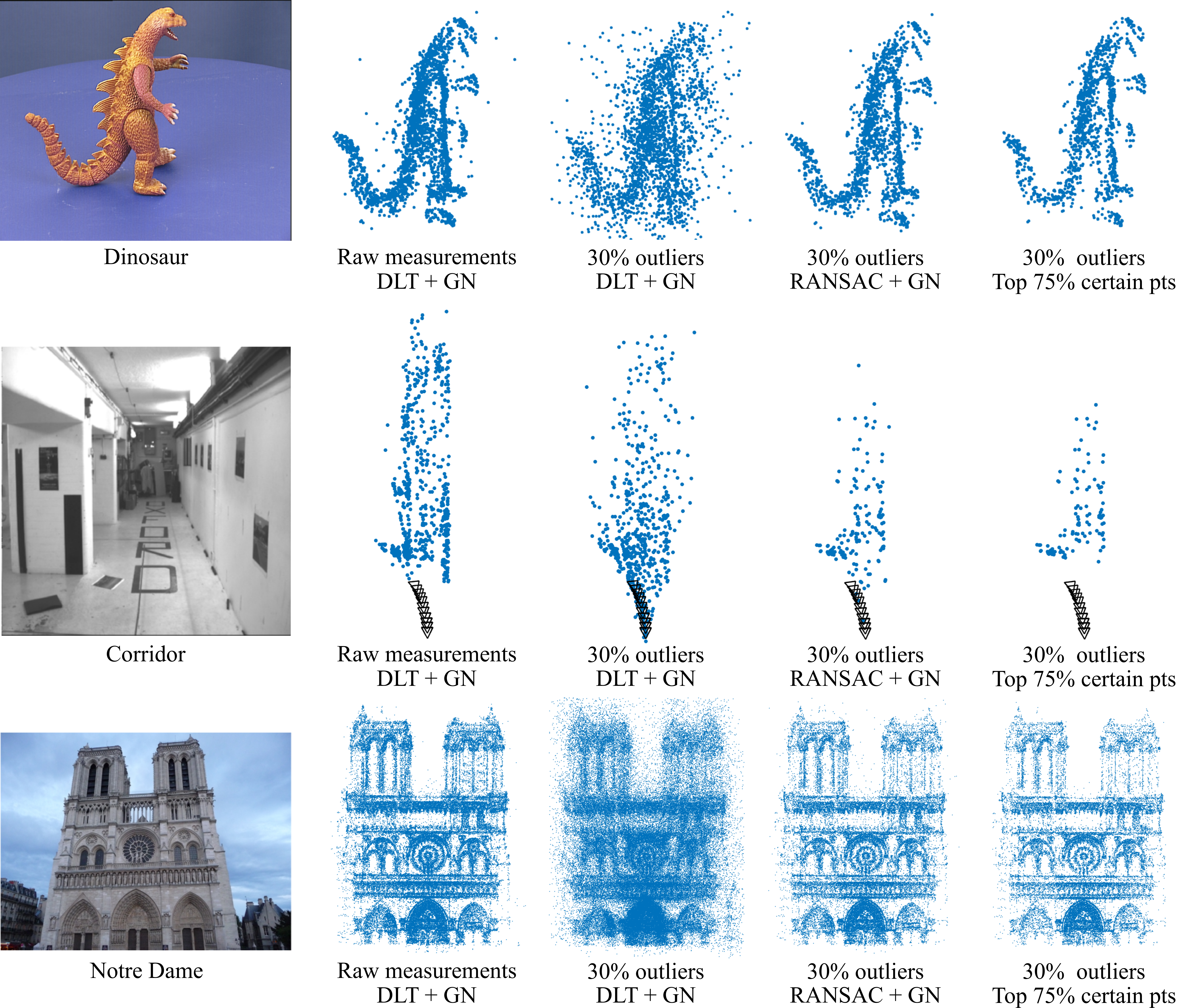}
\caption{
\textbf{1}$^\textbf{st}$ \textbf{column:} Sample images. 
\textbf{2}$^\textbf{nd}$ \textbf{column:} Reconstruction using the \texttt{GN} method (initialized by the \texttt{DLT} method), assuming no outliers.
\textbf{3}$^\textbf{rd}$ \textbf{column:} Reconstruction using the same method on outlier-contaminated measurements. 
We perturb 30\% of the measurements with uniform noise between 10 and 100 pix.
\textbf{4}$^\textbf{th}$ \textbf{column:} Reconstruction using our method (Alg. \ref{al:proposed}) on the same contaminated measurements.
We observe that our RANSAC method is effective against the outliers.
\textbf{5}$^\textbf{th}$ \textbf{column:} From the previous result, we prune the top 25\% of the most uncertain points identified by our method (Sect. \ref{subsec:uncertainty}).
We use the uncertainty model we learned from the simulations in Sect. \ref{subsec:result_uncertainty}.
Notice that some of the most inaccurate points are removed (see the supplementary material for larger images).
}
\label{fig:3D_reconstruction}
\end{figure*}

\subsection{Results on Real Data}
\vspace{-0.5em}
We evaluate our method on three real datasets: Dinosaur \cite{oxford_dataset}, Corridor \cite{oxford_dataset} and Notre Dame \cite{notredame_dataset}.
We only consider the points that are visible in three or more views.
In our algorithm, we use the same parameters as in Sect. \ref{subsec:triangulation_performance} and discard the point that is visible in less than three views after RANSAC.
Fig. \ref{fig:3D_reconstruction} shows the 3D reconstruction results.

\section{Conclusions}
\vspace{-0.5em}
In this work, we presented a robust and efficient method for multiview triangulation and uncertainty estimation.
We proposed several early termination criteria for two-view RANSAC using the midpoint method, and showed that it improves the efficiency when the outlier ratio is high.
We also compared the three local optimization methods (\texttt{DLT}, \texttt{LinLS} and \texttt{GN}), and found that \texttt{DLT} and \texttt{GN} are similar (but better than \texttt{LinLS}) in terms of 3D accuracy, while \texttt{GN} is sometimes much more accurate than \texttt{DLT} in terms of 2D accuracy.
Finally, we proposed a novel method to estimate the uncertainty of a triangulated point based on the number of (inlying) views, the mean reprojection error and the maximum parallax angle.
We showed that the estimated uncertainty can be used to control the 3D accuracy.
An extensive evaluation was performed on both synthetic and real data.

\clearpage
\FloatBarrier

{\small
\balance
\bibliographystyle{ieee}
% \bibliography{egbib}

}

\end{document}